\documentclass[a4paper]{article}

\usepackage{INTERSPEECH2016}

\usepackage{graphicx}
\usepackage{amssymb,amsmath,bm}
\usepackage{textcomp}

 \usepackage{graphicx}
\usepackage{array}
\usepackage{color}
\usepackage{upgreek,amssymb,amsmath}
\usepackage{colortbl}
\definecolor{grey}{rgb}{0.8, 0.8, 0.84}
\definecolor{grey2}{rgb}{0.92, 0.92, 0.97}

\usepackage{subfig}
\usepackage{wrapfig}
\usepackage{hhline}

\newcolumntype{X}[1]{>{\centering\arraybackslash}p{#1}}
\newcolumntype{A}{>{\centering\arraybackslash}p{0.4cm}}
\newcolumntype{C}{>{\centering\arraybackslash}p{0.6cm}}
\newcolumntype{E}{>{\arraybackslash}p{0.7cm}}
\newcolumntype{B}{>{\arraybackslash}p{1.45cm}}
\newcolumntype{Y}{>{\arraybackslash}p{1.42cm}}
\newcolumntype{D}{>{\centering\arraybackslash}p{2.75cm}}

\ninept

\title{Sequential recurrent neural networks for language modeling}

\makeatletter
\def\name#1{\gdef\@name{#1\\}}
\makeatother \name{{\em Youssef Oualil\textsuperscript{1,2}, Clayton Greenberg\textsuperscript{1,2,3}, Mittul Singh\textsuperscript{1,3},  Dietrich Klakow\textsuperscript{1,2,3} 
\thanks{This research was funded by the German Research Foundation (DFG) as part of SFB 1102.}}}

\address{\textsuperscript{1}Spoken Language Systems (LSV)
\\\textsuperscript{2}Collaborative Research Center on Information Density and Linguistic Encoding
\\\textsuperscript{3}Graduate School of Computer Science 
\\Saarland University, Saarbr\"{u}cken, Germany \\
  {\small \tt \{firstname.lastname\}@lsv.uni-saarland.de}
}

\begin{document}

\maketitle
%
\begin{abstract}
Feedforward Neural Network (FNN)-based language models estimate the
probability of the next word based on the history of the last N words, whereas
Recurrent Neural Networks (RNN) perform the same task based only on the last
word and some context information that cycles in the network. This paper
presents a novel approach, which bridges the gap between these two categories of
networks. In particular, we propose an architecture which takes advantage of
the explicit, sequential enumeration of the word history in FNN structure while
enhancing each word representation at the projection layer through recurrent context
information that evolves in the network. The context integration is performed
using an additional word-dependent weight matrix that is also learned during
the training. Extensive experiments conducted on the Penn Treebank (PTB) and
the Large Text Compression Benchmark (LTCB) corpus showed a significant
reduction of the perplexity when compared to state-of-the-art feedforward as
well as recurrent neural network architectures.
\end{abstract}

\section{Introduction}
\label{sec:intro}

A high quality Language Model (LM) is considered to be an integral component of many systems for language technology applications, 
such as speech recognition~\cite{Katz1987}, machine translation~\cite{Brown1990}, etc.  
The goal of an LM is to identify probable sequences of predefined linguistic units, which are typically words. 
Semantic and syntactic properties of the language, encoded by the LM, guide these predictions.

Intrinsically, the performance of an LM can be evaluated based upon its ability to predict the next word given its context. 
The most common approach to build such models is the word count-based method, which is commonly known as $N$-gram language 
modeling~\cite{Rosenfeld2000,KN1995}. By simply enumerating all possibilities over a short span of words and assigning probabilities to them directly, 
$N$-grams were difficult to outperform for a very long time.

The introduction of neural networks for language modeling led to a significant improvement over these standard models. 
This was mainly due to the continuous word representations they provide, which typically overcome the exponential growth 
of parameters that $N$-gram models require to enumerate possibilities. Bengio et al.~\cite{Bengio2003} proposed a Feedforward 
Neural Network (FNN) for language modeling, as an alternative to $N$-grams, to estimate the probability of a given word sequence 
while considering a fixed context (word history) size. This approach was very successful and has been shown to outperform a 
mixture of different other models~\cite{Goodman2001b}, and to significantly improve speech recognition performance~\cite{Schwenk2005}. 

In order to overcome the fixed context size constraint and to capture long range dependencies known to be present in language, 
Mikolov et al.~\cite{Mikolov2010,Mikolov2011} proposed a Recurrent Neural Network (RNN) which allows context information to cycle in the network. Another recurrence-based network architecture, Long-Short Term Memory (LSTM)~\cite{Sundermeyer12}, addresses some learning issues from the original RNN and explicitly controls the longevity of context information in the network.  

Contrary to FNN, recurrent models such as RNN and LSTM predict the next word based only on the current word and the context representation.  Therefore, they lose information about word position rather quickly and cannot model short range dependencies as well as FNN and $N$-grams. For example, English has position-dependent patterns such as ``he $*$ he'' (``he said he'', ``he mentioned he'', \ldots).  The position of ``he'' is essential for making the right prediction in this case, and the recurrent models are not designed to encode that.  Rather, they are better for smooth incremental updates and hence for longer range dependencies.

This paper proposes a novel approach that models short range dependencies like FNN and long range dependencies like RNN.  
In particular, the hidden layers combine explicit encoding of the local context and a recurrent architecture, which allows the 
context information to sequentially evolve in the network at the projection layer. In the first step, the word representation are enhanced using the context information. This step maps the word representations from a universal embedding space into a context-based space. 
Then, the system performs the next word prediction as it is typically done in FNN. The learning of the network weights uses the Back-Propagation 
Through Time (BPTT) algorithm similarly to RNN. The main difference here is the additional network error resulting from the additional 
sequential connections. This paper also shows that learning of word-dependent sequential connections can substantially improve the 
performance of the proposed network. 

We proceed as follows. Section~\ref{sec:NN} presents a brief overview of FNN and RNN models. Section~\ref{sec:SRNN} introduces the proposed architecture which combines these two models. Then, Section~\ref{sec:EXP} evaluates the proposed network in comparison to different state-of-the-art language models for perplexity on the PTB and the LTCB corpus. Finally, we conclude in Section~\ref{sec:CC}.

\section{Neural Network Language Models}
\label{sec:NN}
The goal of a language model is to estimate the probability distribution $p(w_1^T)$ of word sequences $w_1^T = w_1,\cdots,w_T$.
Using the chain rule, this distribution can be expressed as
\vspace{-2mm}
\begin{equation}
  \label{eq:prob}
\displaystyle{ p(w_1^T) = \prod_{t=1}^T{p(w_t|w_1^{t-1})} }
\vspace{-1mm}
\end{equation}
The rest of this section shows how FNN and RNN are used to approximate this probability distribution.

\subsection{Feedforward Neural Networks}
\label{ssec:FNN}
Similarly to $N$-gram models, FNN uses the Markov assumption of order N-1
to approximate~(\ref{eq:prob}) according to
\vspace{-1mm}
\begin{equation}
  \label{eq:fnn-app-prob}
	 \displaystyle{ p(w_1^T) \approx \prod_{t=1}^T{p(w_t|w_{t-N+1}^{t-1})} }
	\vspace{-1mm}
\end{equation}
Subsequently, each of the terms involved in this product, i.e, $p(w_t|w_{t-N+1}^{t-1})$, 
is estimated, separately, in a single bottom-up evaluation of the network according to 
\vspace{-0.5mm}
\begin{align}
\label{eqn:eqfnn-1}
 P_{t-i} &=  X_{t-i} \cdot U \hspace{1mm}, \quad \quad i=N-1,\cdots,1  \\  \label{eqn:eqfnn-2}
 H_{t}   &= f \left( \sum_{i=1}^{N-1} P_{t-i} \cdot V_i \right) \\ \label{eqn:eqfnn-3}
 O_t   &= g \left( H_{t} \cdot W\right)
  \vspace{-1mm}
 \end{align}
$X_{t-i}$ is a one-hot encoding of the word $w_{t-i}$, whereas 
the rows of $U$ encode the continuous word representations (i.e, embeddings).
Thus, $P_{t-i}$ is the continuous representation of the word $w_{t-i}$.
$W$ and $V=[V_1,\cdots,V_{N-1}]$ are the network connection weights, which
are learned during training in addition to $U$. Moreover, $f(\cdot)$ is an activation 
function, whereas $g(\cdot)$ is the softmax function. Figure~(\ref{fig:fnn}) 
shows an example of an FNN with a fixed context size $N-1=3$ with a single hidden layer. 

\subsection{Recurrent Neural Networks}
\label{ssec:RNN}
An RNN attempts to capture the complete history in a context vector 
$h_{t}$, which represents the state of the network and evolves in time.
Therefore, it approximates~(\ref{eq:prob}) according to
%
\begin{equation}
  \label{eq:rnn-app-prob}
	 \displaystyle{ p(w_1^T) \approx \prod_{t=1}^T{p(w_t|w_{t-1}, h_{t-1})} = \prod_{t=1}^T{p(w_t|h_{t})} }
\vspace{-1mm}
\end{equation}
RNN evaluates this distribution similarly to FNN. The main difference 
occurs in Equations~(\ref{eqn:eqfnn-1}) and~(\ref{eqn:eqfnn-2}) which are combined into

\vspace{-1mm}
\begin{equation}
  \label{eq:eqrnn-1}
	  H_{t} = f \left( X_{t-1} \cdot U + H_{t-1} \cdot V \right)
	  \vspace{-3mm}
\end{equation}
\vspace{-3mm}
\begin{figure}[!h]
 \centering
   \subfloat[FNN]{\label{fig:fnn}\includegraphics[totalheight=0.14\textheight, width=0.25\textwidth]{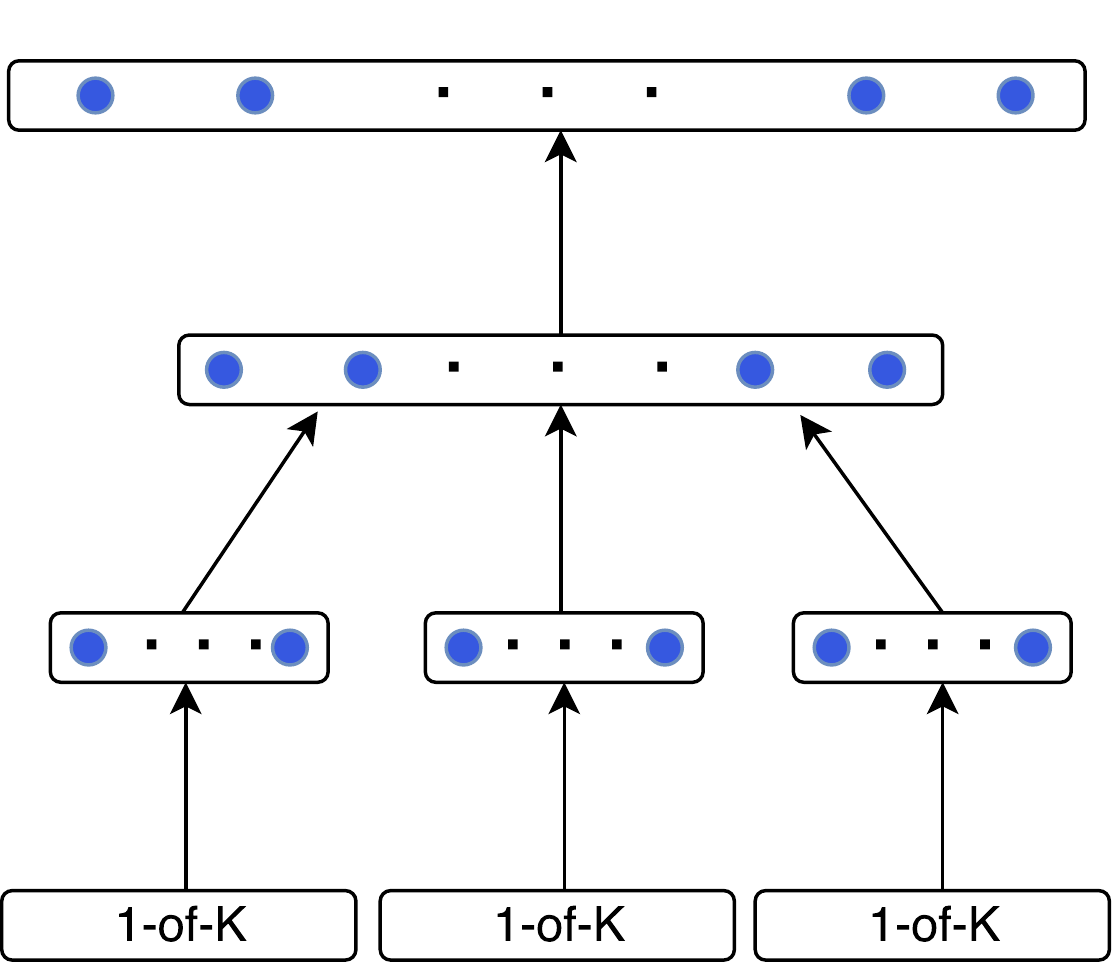}}
   \subfloat[RNN]{\label{fig:rnn}\includegraphics[totalheight=0.14\textheight,width=0.21\textwidth]{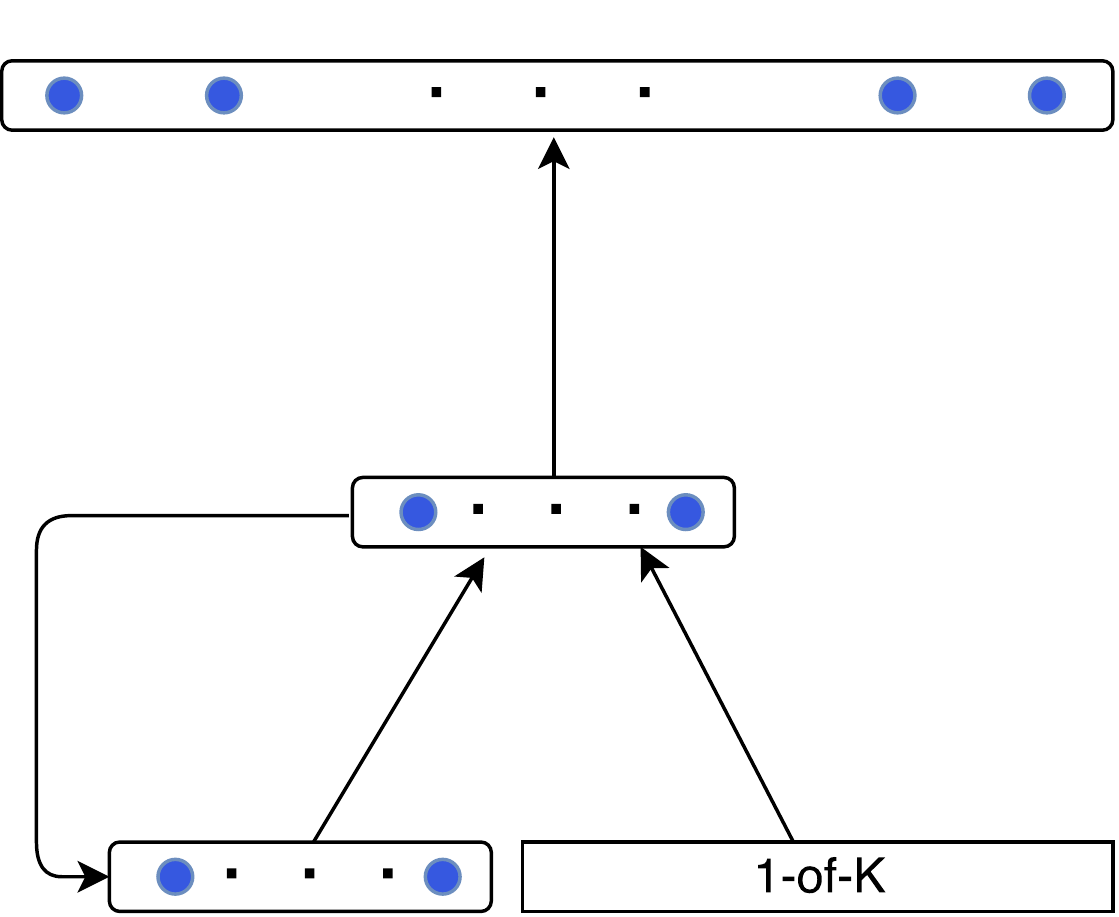}}
   \vspace{-2mm}
 \caption{\it FNN vs RNN Architecture.} 
 \vspace{-1mm}
 \label{fig:MSL}
\end{figure}

Figure~(\ref{fig:rnn}) shows an example of a standard RNN.
The next Section will show how an RNN can be extended to explicitly 
model short range dependencies through additional sequential connections.

%
\section{Sequential Recurrent Neural Network}
\label{sec:SRNN}
The main difference between an RNN and an FNN is the context representation. More precisely, 
The context layer $H_{t}$ of an FNN is estimated based on a fixed context size i.e, the last $N-1$ words,
whereas in an RNN, $H_{t}$ is constantly updated (at each time iteration) using only the last word and context at time $t-1$.  

\subsection{The proposed Neural Architecture}
\label{ssec:Archi-SRNN}

\begin{figure}[!ht]
\vspace{-0.1mm}
  \centering
  \includegraphics[trim=70 00 50 5 ,clip, totalheight=0.16\textheight, width=1.0\linewidth]{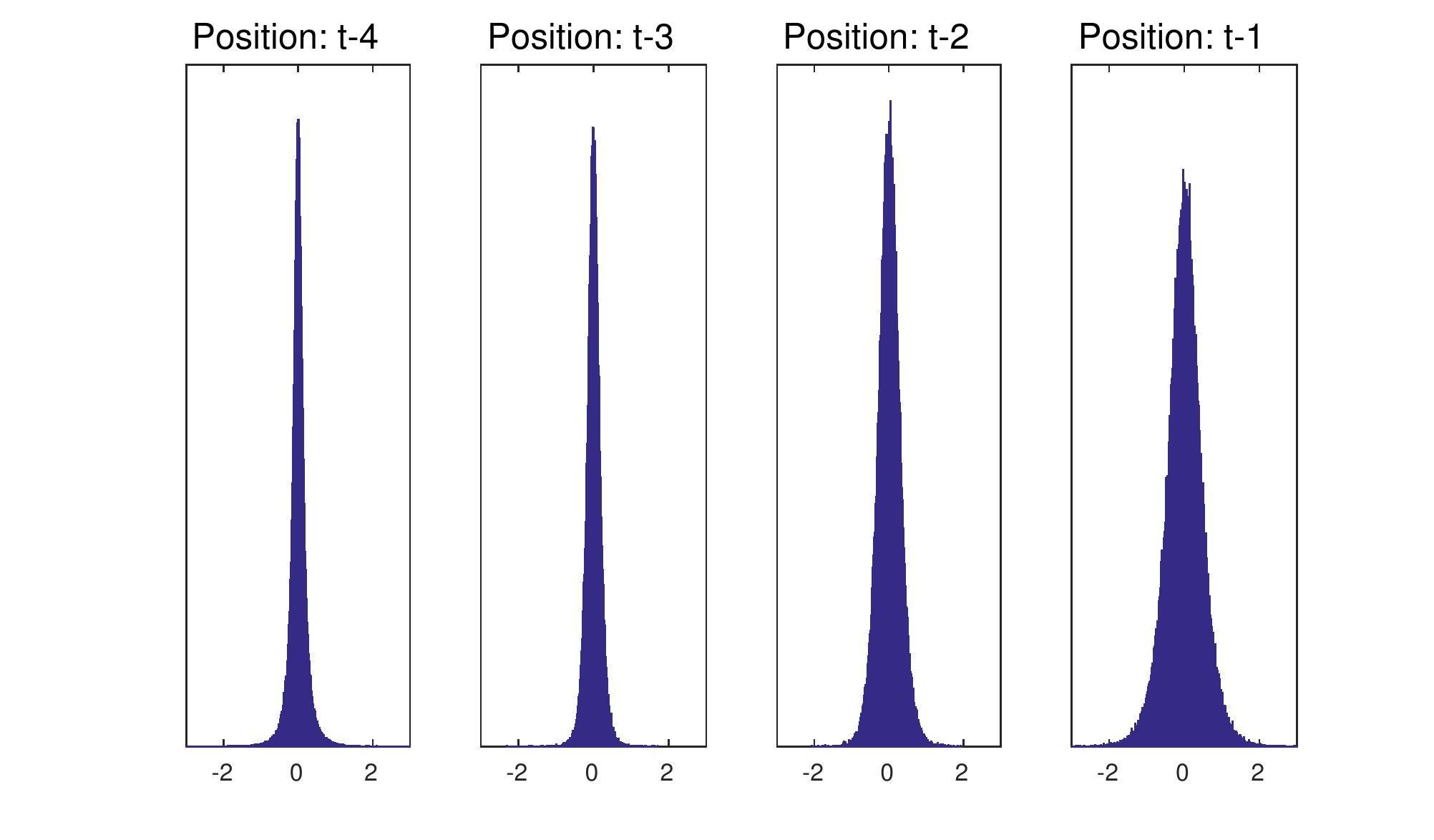}
   \vspace{-8mm}
	\caption{Histograms of the projection-to-hidden weights $V_1$,$V_2$,$V_3$ and $V_4$ (see Figure~\ref{fig:srnn}) for each of the 4 word positions of an SRNN (N=5) trained on LTCB.
	These histograms show that the magnitude of the weights decays with the word position (from $t-1$ to $t-4$) but does not nullify. Thus, the farther word positions still
        capture some residual/additional context.}
  \label{fig:pos-hist}
\end{figure}

We propose in this paper an architecture which captures short range dependencies 
over the last $N-1$ word positions as it is done in FNN, and the long range context through 
recurrence, similarly to RNN. 
%
%
The design of this structure is motivated by the inefficiency of 
RNN to model position dependent patterns, which 
are particularly frequent in conversational speech. RNN loses information about 
word position quickly and therefore cannot efficiently model short range dependencies. FNN and N-gram models,
however, are designed as position-dependent models, which deal only with short-term context. 
%
%
Extending RNN structure to explicitly represent the short term history as it is done in FNN will 1) help improve 
the modeling of short range context, as it will 2) allow the network to capture any residual/additional context 
information that may be present in the past $i=t-N+1, \cdots, t-2$ time iterations but 
which may have been lost during the last context update, which is based only on the last word  at $t-1$ (See illustration in Figure~\ref{fig:pos-hist}). 
In the worst case scenario, the context information will be simply redundant and is 
expected not to harm the performance. The rest of this Section introduces the 
mathematical formulation of this approach. 
%
%

The proposed Sequential Recurrent Neural Network (SRNN) approximates~(\ref{eq:prob}) according to
\vspace{-1mm}
\begin{equation}
  \label{eq:srnn-app-prob}
	\displaystyle{ \!\!  p(w_1^T) \! \approx \!\!  \prod_{t=1}^T{p(w_t|w_{t-N+1}^{t-1}, h_{t-N+1})} 
	                        \! = \! \prod_{t=1}^T{p(w_t|h_{t-N+2}^{t})} }  \!\!\!  
\end{equation}
The proposed architecture to estimate~(\ref{eq:srnn-app-prob}) explicitly represents the history
over the last $N-1$ word positions as it is done in FNN to approximate~(\ref{eq:fnn-app-prob}) while it enhances the actual word representations 
using the recurrent context information, which propagates sequentially within the network. 
Furthermore, restricting the context to a 1-word history window (N=2) in~(\ref{eq:srnn-app-prob}) leads to the RNN approximation in~(\ref{eq:rnn-app-prob}). 
Therefore, the proposed approach can be seen as an extension of the standard RNN to explicitly
model and capture short range context.

The additional sequential connections allow the context information to propagate from the past to the future within the 
network. These connections can be defined as a Word-Independent (WI) recurrence vector, which fixes the amount of context 
information allowed to propagate in the network, as they can be designed as Word-Dependent (WD) vectors. In this case, 
each word will have its own context weight vector, which will typically learn which context ``neurons'' are relevant 
for that particular word and therefore scales each context unit accordingly.  

\vspace{-3mm}
\begin{figure}[!ht]
  \centering
  \includegraphics[totalheight=0.16\textheight, width=0.98\linewidth]{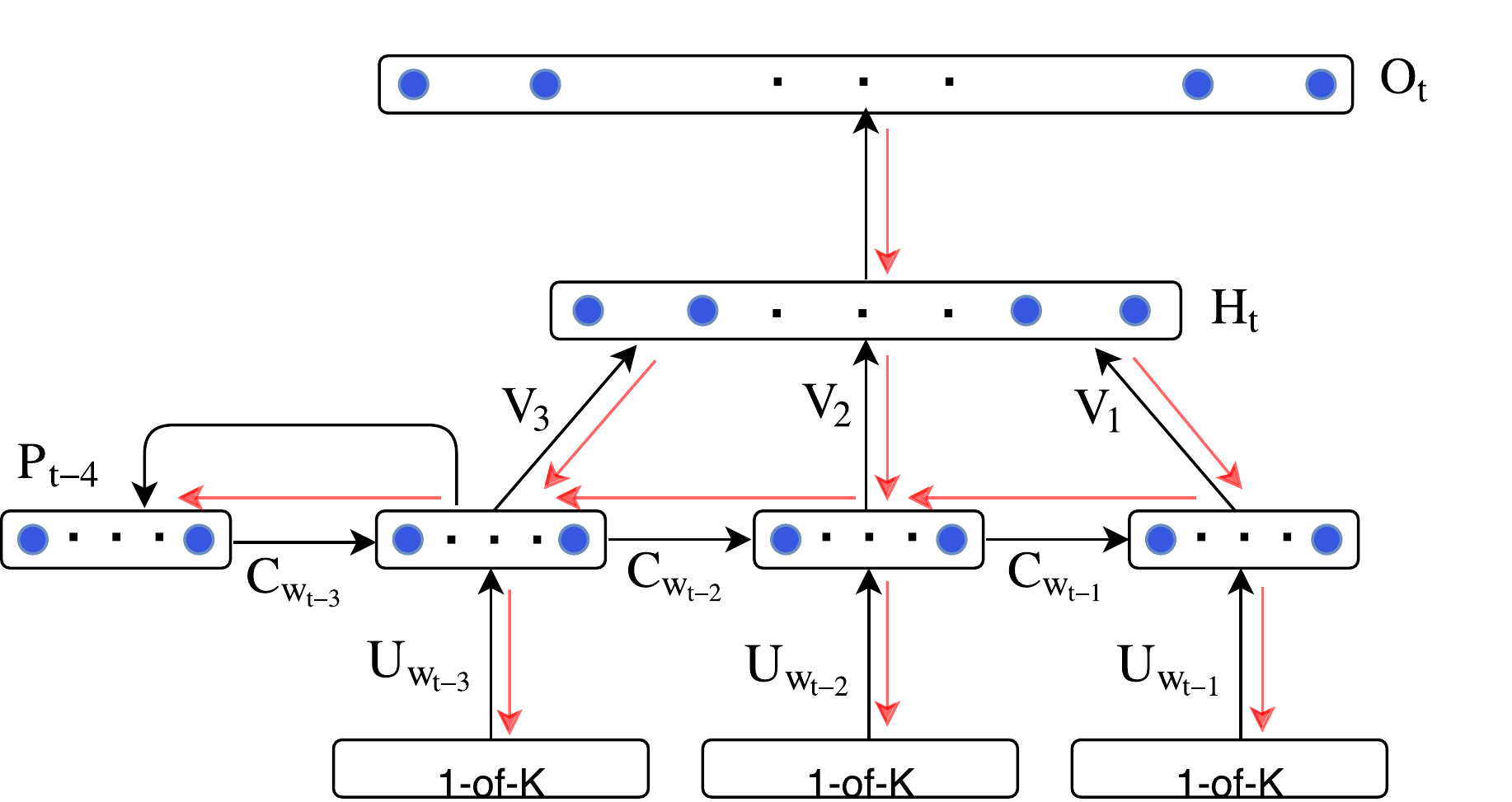}
  \vspace{-1mm}
	\caption{Sequential Recurrent Neural Network architecture. The backward path (red arrows) 
	shows the error propagation during training (this figure does not include BPTT).}
  \label{fig:srnn}
\end{figure}
\vspace{-2mm}

The network evaluation is performed similarly to FNN, the main difference 
occurs in Equation~(\ref{eqn:eqfnn-1}), which becomes in the case of the word-indepdent model
%
\begin{equation}
  \label{eq:eqsrnn-independent}
 P_{t-i} = f_s(X_{t-i} \cdot U + C \odot P_{t-i-1}) , \quad i=N-1,\cdots,1 
\end{equation}
as it becomes in the case of the word-dependent model
\begin{equation}
  \label{eq:eqsrnn-dependent}
 \! P_{t-i}\! = \! f_s(X_{t-i} \cdot U + C_{w_{t-i}} \odot  P_{t-i-1}), \hspace{1mm} i=N-1,\cdots,1  \! \! \! \!
\end{equation}

where $f_s(\cdot)$ is an activation function and $\odot$ is the element-wise product operator. 
$C$ is the word-independent recurrence weight vector, whereas $C_{w_{t-i}}$ is the word-dependent context weight corresponding to the word $w_{t-i}$.
Figure~(\ref{fig:srnn}) shows an example of an SRNN with three additional sequential connections ($N-1=3$) and a single hidden layer.

The proposed SRNN model is a general architecture that includes different networks. In particular, 
setting $C\!=\![0,\cdots,0]$ and $f_s(x)\!=\!x$ results in the classical FNN architecture,
whereas setting $N=2$ leads to a standard RNN with a diagonal recurrence matrix and an additional non-recurrent layer. 
Moreover, setting $C$ to a fixed value in $[0,1]$ and $f_s(x)\!=\!x$ leads to the Fixed-size Ordinally-Forgetting Encoding 
(FOFE)~\cite{FOFE2015} architecture, which was proposed to uniquely encode word sequences. 

The proposed model replaces the universal word embeddings at the projection layer of 
an FNN by context-dependent word embeddings. More particularly, both Equations~(\ref{eq:eqsrnn-independent}) 
and (\ref{eq:eqsrnn-dependent}) show that each word representation is enhanced using the context 
information before proceeding to the next word prediction. Therefore, we can see this particular step as 
a transformation from the universal embedding space into a context-dependent space with a better 
discrimination of words.

\subsection{SRNN Training}
\label{ssec:Train-SRNN}
The parameters to train for an SRNN are the word embeddings $U$, the 
project-to-hidden connection weights $V=[V_1,\cdots,V_{N-1}]$, the hidden-to-output
connection weights $W$ and the context weight vector $C$ for the WI model, 
or $C=[C_1^\intercal,\cdots,C_K^\intercal]^\intercal$ ($K$ is the vocabulary size) for the WD model.  
In this case, each word $w$ in the vocabulary will be characterized by two learnable vectors, namely, the continuous 
representation (embedding) $U_w$ and the context weight $C_w$. 

Similarly to RNN, the parameter learning of an SRNN architecture follows the standard Back-Propagation 
Through Time (BPTT) algorithm. The main difference occurs at the projection layer, where the additional error 
vectors resulting from the sequential connections should be taken into account (See example or error propagation in 
Figure~\ref{fig:srnn}) before unfolding the network in time. 

\vspace{-2mm}

\section{Experiments and Results}
\label{sec:EXP}
\subsection{Experimental Setup}
We evaluated the proposed architecture on two different benchmark tasks. 
The first set of experiments was conducted on the Penn Treebank (PTB) corpus using the standard division, e.g.~\cite{Mikolov2011,FOFE2015}: sections 0-20 are used for training
while sections 21-22 and 23-24 are used for validation and testing. The vocabulary was limited to the most 10k frequent words while the remaining
words were all mapped to the token $<$unk$>$. In order to evaluate how the proposed
approach scales to large corpora, we run a set of experiments on
the Large Text Compression Benchmark (LTCB)~\cite{Mahoney2011}.
This corpus is based on the enwik9 dataset which contains the first $10^9$ 
bytes of enwiki-20060303-pages-articles.xml. We adopted the same 
training-test-validation data split and preprocessing from~\cite{FOFE2015}. All but the 80k most frequent words were replaced by $<$unk$>$.
Details about the sizes of these two corpora and the percentage of 
Out-Of-Vocabulary (OOV) words that were mapped to $<$unk$>$ can be found in Table~\ref{tab:corpora}.
\vspace{-2mm}
\renewcommand{\tabcolsep}{2.2pt}
\begin{table}[!th]
  \caption{\label{tab:corpora} {\it Corpus size in number of words and $<$unk$>$ rate.}}
  \vspace{-2mm}
  \centerline{
  \begin{tabular}{| c || c | c || c | c || c | c | }
      \hline
  \multicolumn{1}{|c||}{} & \multicolumn{2}{c||}{Train} & \multicolumn{2}{c||}{Dev} & \multicolumn{2}{c|}{Test} \\
    \hline
    Corpus & \#W & $<$unk$>$ & \#W & $<$unk$>$ & \#W & $<$unk$>$ \\
    \hline
    \hline
    PTB  & 930K & 6.52\% & 82K & 6.47\% & 74K & 7.45\% \\
    \hline
    LTCB & 133M & 1.43\% & 7.8M & 2.15\% & 7.9M & 2.30\%  \\
    \hline
  \end{tabular}}
\end{table}
\vspace{-2mm}

The proposed approach (SRNN) is compared to different systems including the $N$-gram 
Kneser-Ney (KN) model and different feedforward and recurrent neural architectures. 
For feedforward networks, the baseline systems include 1) the FNN-based LM~\cite{Bengio2003}
as well as the 2) Fixed-size Ordinally Forgetting Encoding (FOFE) approach, which was implemented as a 
feedforward sentence-based model~\cite{FOFE2015}. 
The FOFE results were obtained using the FOFE toolkit~\cite{FOFE2015}. 
The results are reported for different context sizes (N-1=1,2 and 4) and different numbers of hidden layers (1 or 2).
Regarding recurrent models, we compare the proposed approach to 3) the full RNN 
(without classes)~\cite{Mikolov2011}, 4) to a deep RNN~\cite{Pascanu2013}, which investigates different ways of 
adding hidden layers to RNN, and finally 5) to the LSTM architecture~\cite{Sundermeyer12}, which explicitly
regulates the amount of information that propagates in the network. 
%
\subsection{PTB Experiments}

For the PTB experiments, the FNN, FOFE and SRNN architectures have similar configurations. That is, 
the hidden layer(s) size is 400 with all hidden units using the Rectified Linear Unit (ReLu) i.e., $f(x)=max(0,x)$,
as an activation function, whereas the word representation (embedding) size was set to 200 for FNN, FOFE and LSTM and 100 for SRNN.
The latter uses $f_s=tanh(\cdot)$ as sequential activation function.  The hidden layer size of RNN and LSTM were set to 
400 and follow the original configuration proposed in~\cite{Mikolov2011} and~\cite{Sundermeyer12}, respectively.
We also use the same learning setup adopted in~\cite{FOFE2015}. Namely, 
we use the stochastic gradient descent algorithm with a mini-batch size of 200, 
the learning rate is initialized to 0.4, the momentum is set to 0.9, the weight 
decay is fixed to $4.10^{-5}$ and the training is done in epochs. The weights initialization follows the 
normalized initialization proposed in~\cite{Glorot2010}. Similarly to~\cite{Mikolov2010}, 
the learning rate is halved when no significant improvement in the log-likelihood of the validation data is
observed. Then, we continue with seven more epochs while halving the learning rate after each epoch. The BPTT was set to 5 time steps.
In the tables below, WI-SRNN refers to the word-independent SRNN model proposed in~(\ref{eq:eqsrnn-independent}), whereas 
WD-SRNN refers to the word-dependent model in~(\ref{eq:eqsrnn-dependent}). For both models, the context connection weights, $C$, 
were randomly initialized in $[0,1]$. In order to compare to the FOFE approach, we also report results where $C$ is reduced to 
a scalar forgetting factor that is fixed at 0.7. This is denoted as WI-SRNN$^*$ in the tables below. We report the results 
in terms of perplexity (PPL), Number of model Parameters (NoP) and the training speed, 
which is defined as the number of words processed per second (w/s) on a GTX TITAN X GPU.
\vspace{-1mm}
\renewcommand{\tabcolsep}{2.4pt}
\begin{table}[!h]
  \caption{\label{tab:ptb} {\it LMs performance on the PTB test set.}}
  \vspace{-2mm}
    \centerline{
  \begin{tabular}{| c | c | c | c || c | c | c || c || c |}
     \hhline{~|---||---||-||-}
     \multicolumn{1}{c|}{} & \multicolumn{3}{>{\columncolor{grey}}c||}{ model } &  \multicolumn{3}{>{\columncolor{grey}}c||}{ model+KN5 } 
     & \multicolumn{1}{>{\columncolor{grey}}c||}{ NoP } & \multicolumn{1}{>{\columncolor{grey}}c|}{ w/s } \\
     \hline
      N-1= & 1 & 2 & 4 & 1 & 2 & 4 & 4 & 4 \\
     \hline
      & \multicolumn{8}{>{\columncolor{grey2}}c|}{ 1 Hidden Layer }  \\
      \hline
      FNN          & 176 & 131 & 119  & 132 & 116 & 107 & 6.32M & 24.3K \\
      FOFE         & 123 & 111 & 112  & 108 & 100 & 101 & 6.32M & 17.2K \\
      \hline
      WI-SRNN$^*$  & 117 & 110 & 109  & 105 & 100 &  99  & 5.16M & 12.9K \\ 
      WI-SRNN      & 112 & 107 & 107  & 102 &  98 &  97  & 5.16M & 11.2K \\ 
      WD-SRNN      & 109 & 106 & 106  &  99 &  96 &  95  & 6.16M & 10.4K \\ 
     \hline
      & \multicolumn{8}{>{\columncolor{grey2}}c|}{ 2 Hidden Layers }  \\
      \hline
      FNN          & 176 & 129 & 114  & 132  & 114 & 102 & 6.48M & 21.8K \\
      FOFE         & 116 & 108 & 109  & 104  & 98  &  97 & 6.48M & 16.6K \\
      \hline
      WI-SRNN$^*$ & 114 & 108 & 107  & 102  & 98  &  96  & 5.32M  & 10.8K \\ 
      WI-SRNN     & 109 & 105 & 104  &  99  & 96  &  94  & 5.32M  & 9.6K \\ 
      WD-SRNN     & 108 & 103 & 104  &  97  & 94  &  94  & 6.32M  & 9.2K \\ 
     \hline
      & \multicolumn{8}{>{\columncolor{grey2}}c|}{ Recurrent Models }  \\
      \hline
      RNN      & \multicolumn{3}{c||}{ 123 }  & \multicolumn{3}{c|}{ 107 } & 8.16M & 20.6K \\
      \hline
      Deep RNN & \multicolumn{3}{c||}{107.5} & \multicolumn{3}{c|}{ --- } & 6.96M & ---  \\
      \hline
      LSTM     & \multicolumn{3}{c||}{ 114 } & \multicolumn{3}{c|}{ 99 } & 6.96M & 7.6K \\
      \hline
  \end{tabular} }
\end{table}

Table~\ref{tab:ptb} shows the LMs evaluation on the PTB test set. We can clearly see that 
the proposed approach outperforms all other models using the lowest Number of model 
Parameters (NoP) among all configurations. This also includes other models that were reported in the literature, 
such as RNN with maximum entropy~\cite{Mikolov2011b}, random forest LM~\cite{Xu2007}, 
structured LM~\cite{Filimonov2009} and syntactic neural network LM~\cite{Emami2004}. More particularly, 
SRNN with two hidden layers achieves a comparable performance to a mixture of RNNs~\cite{Mikolov2011c}. 
We can also conclude that the explicit modeling of short range dependencies through sequential connections improves the performance. More precisely,
the results show that increasing the history window (1, 2 and 4)  improves the performance for all SRNN models.
Table~\ref{tab:ptb} also shows that using a fixed scalar forgetting factor (WI-SRNN$^*$) leads to a slight improvement 
over the FOFE approach, which is mainly due to the additional non-linear activation function $f_s$.
Furthermore, the word-dependent (WD-SRNN) model slightly outperforms the word-independent model (WI-SRNN) but 
with a non-negligible increase in the number of parameters. Regarding the training speed, we can conclude that
training an SRNN model requires approximately twice the time needed for FFN and RNN, whereas it needs less time compared to LSTM. 
\subsection{LTCB Experiments}
			
The LTCB experiments use the same PTB setup with minor changes. 
The results shown in Table~\ref{tab:ltcb} follow the same experimental setup used in~\cite{FOFE2015}. More precisely, 
these results were obtained without usage of momentum or weight decay whereas the mini-batch size was set to 400.
The FNN and FOFE architectures contain 2 hidden layers of size 600 (or 400) whereas RNN and SRNN have 
a single hidden layer of size 600. In order to compare to ~\cite{FOFE2015}, the forgetting factor C of WI-SRNN$^*$ is fixed at 0.6.
\vspace{-1mm}
\renewcommand{\tabcolsep}{2.7pt}
\begin{table}[!h]
\caption{\label{tab:ltcb} {\it LMs Perplexity on the LTCB test set.}}
\vspace{-2mm}
\centerline{
\begin{tabular}{| c | c | c | c || c |}
\hhline{~---||-}
\multicolumn{1}{c|}{} & \multicolumn{3}{>{\columncolor{grey}}c||}{model} & \multicolumn{1}{>{\columncolor{grey}}c|}{NoP} \\
\hline
Context Size M=N-1 & 1 & 2 & 4 & 4  \\
\hline
KN  &  239  &  156  & 132  & ---  \\
\hline
FNN  [M*200]-600-600-80k & 235 & 150 & 114 & 64.84M \\
FOFE [M*200]-400-400-80k & 120 & 115 & 108 & 48.48M \\
FOFE [M*200]-600-600-80k & 112 & 107 & 100 & 64.84M \\
\hline
\hline
WI-SRNN$^*$ [M*200]-600-80k & 110  & 102  & 94 & 64.48M \\
WI-SRNN [M*200]-600-80k     &  85  &  80  & 77 & 64.48M \\
WD-SRNN [M*200]-600-80k     &  77  &  74  & 72 & 80.48M \\
\hline
\hline
RNN [600]-600-80k  & \multicolumn{3}{c||}{ 85 } & 96.36M   \\
\hline
\end{tabular} }
\end{table}
\vspace{-3mm}

The LTCB results shown in Table~\ref{tab:ltcb} generally confirm 
the PTB conclusions. In particular, we can see that SRNN models outperform all other models 
while requiring comparable or fewer model parameters. Moreover, the WI-SRNN$^*$ model with a 
single hidden layer slightly outperforms FOFE (2 hidden layers). 
These results, however, show a more significant improvement 
for the WD-SRNN model and for the increased window size (from 1 to 4)
compared to the improvement obtained on the PTB. This is mainly due to the large 
amount of LTCB training data, which allows us to train richer WD 
context vectors.
\renewcommand{\tabcolsep}{2.5pt}
\begin{table}[!h]
\caption{\label{tab:cs2} {\it Examples of top 5 similar words.}}
\vspace{-2mm}
\begin{tabular}{| c | c | c | c | c | c |}
\hline
\multicolumn{2}{|c|}{in} & \multicolumn{2}{c|}{strictly} & \multicolumn{2}{c|}{germany}   \\
\hline
$U_w$  & $ C_w $ &  $U_w$ & $ C_w $ &  $U_w$ & $ C_w $  \\
\hline        
into        &  at    &  solely       & purely      &  italy      &  japan       \\ 
throughout  &  on    &  rigidly      & totally     &  france     &  russia      \\ 
through     &  for   &  broadly      & physically  &  britain    &  italy       \\ 
during      &  their &  purely       & solely      & switzerland &  france      \\ 
within      &  to    &  ostensibly   & technically &  england    &  spain       \\ 
\hline
\end{tabular} 
\end{table}
\vspace{-2mm}

Table~\ref{tab:cs2} shows some word examples with their top 5 cosine similarities for word embeddings
$U_w$ and Euclidean distance for context weights $C_w$. These examples show a general trend, not valid for every example, that the embeddings capture 
semantic (conceptual) similarities and the context weights model syntactic (functional) similarities.

\section{Conclusion and Future Work}
\label{sec:CC}

We have presented a sequential recurrent neural network which 
captures short range dependencies using short history windows, 
and models long range context through recurrent connections. 
Experiments on PTB and LTCB corpora have shown that this architecture substantially outperforms many state-of-the-art neural systems, due to its successful combination of the motivating features of its feedforward and recurrent predecessors.  
Further gains could be made by more optimally controlling the amount of information evolving in the network, as it is done in LSTM, and by more thoroughly addressing long range dependencies. These will be investigated in future work.



\newpage
\eightpt
\bibliographystyle{IEEEtran}
\bibliography{refs_final}

\begin{thebibliography}{10}
\providecommand{\url}[1]{#1}
\csname url@samestyle\endcsname
\providecommand{\newblock}{\relax}
\providecommand{\bibinfo}[2]{#2}
\providecommand{\BIBentrySTDinterwordspacing}{\spaceskip=0pt\relax}
\providecommand{\BIBentryALTinterwordstretchfactor}{4}
\providecommand{\BIBentryALTinterwordspacing}{\spaceskip=\fontdimen2\font plus
\BIBentryALTinterwordstretchfactor\fontdimen3\font minus
  \fontdimen4\font\relax}
\providecommand{\BIBforeignlanguage}[2]{{%
\expandafter\ifx\csname l@#1\endcsname\relax
\typeout{** WARNING: IEEEtran.bst: No hyphenation pattern has been}%
\typeout{** loaded for the language `#1'. Using the pattern for}%
\typeout{** the default language instead.}%
\else
\language=\csname l@#1\endcsname
\fi
#2}}
\providecommand{\BIBdecl}{\relax}
\BIBdecl

\bibitem{Katz1987}
S.~Katz, ``Estimation of probabilities from sparse data for the language model
  component of a speech recognizer,'' \emph{IEEE Transactions on Acoustics,
  Speech, and Signal Processing}, vol.~35, no.~3, pp. 400--401, Mar. 1987.

\bibitem{Brown1990}
P.~F. Brown, J.~Cocke, S.~A.~D. Pietra, V.~J.~D. Pietra, F.~Jelinek, J.~D.
  Lafferty, R.~L. Mercer, and P.~S. Roossin, ``A statistical approach to
  machine translation,'' \emph{Comput. Linguist.}, vol.~16, no.~2, pp. 79--85,
  Jun. 1990.

\bibitem{Rosenfeld2000}
R.~Rosenfeld, ``Two decades of statistical language modeling: Where do we go
  from here?'' in \emph{Proceedings of the IEEE}, vol.~88, 2000, pp.
  1270--1278.

\bibitem{KN1995}
R.~Kneser and H.~Ney, ``Improved backing-off for m-gram language modeling,'' in
  \emph{{IEEE} International Conference on Acoustics, Speech, and Signal
  Processing (ICASSP)}, Detroit, Michigan, USA, May 1995, pp. 181--184.

\bibitem{Bengio2003}
Y.~Bengio, R.~Ducharme, P.~Vincent, and C.~Jauvin, ``A neural probabilistic
  language model,'' \emph{J. Mach. Learn. Res.}, vol.~3, pp. 1137--1155, Mar.
  2003.

\bibitem{Goodman2001b}
J.~Goodman, ``A bit of progress in language modeling, extended version,''
  Microsoft Research, Tech. Rep. MSR-TR-2001-72, 2001.

\bibitem{Schwenk2005}
H.~Schwenk and J.~Gauvain, ``Training neural network language models on very
  large corpora,'' in \emph{Human Language Technology Conference and Conference
  on Empirical Methods in Natural Language Processing (EMNLP)}, Oct. 2005, pp.
  201--208.

\bibitem{Mikolov2010}
T.~Mikolov, M.~Karafi{\'{a}}t, L.~Burget, J.~Cernock{\'{y}}, and S.~Khudanpur,
  ``Recurrent neural network based language model,'' in \emph{11th Annual
  Conference of the International Speech Communication Association
  (INTERSPEECH)}, Makuhari, Chiba, Japan, Sep. 2010, pp. 1045--1048.

\bibitem{Mikolov2011}
T.~Mikolov, S.~Kombrink, L.~Burget, J.~Černocký, and S.~Khudanpur,
  ``Extensions of recurrent neural network language model,'' in \emph{{IEEE}
  International Conference on Acoustics, Speech, and Signal Processing
  (ICASSP)}, May 2011, pp. 5528--5531.

\bibitem{Sundermeyer12}
M.~Sundermeyer, R.~Schl{\"u}ter, and H.~Ney, ``{LSTM} neural networks for
  language modeling,'' in \emph{13th Annual Conference of the International
  Speech Communication Association (INTERSPEECH)}, Portland, OR, USA, Sep.
  2012, pp. 194--197.

\bibitem{FOFE2015}
S.~Zhang, H.~Jiang, M.~Xu, J.~Hou, and L.~Dai, ``The fixed-size
  ordinally-forgetting encoding method for neural network language models,'' in
  \emph{53rd Annual Meeting of the Association for Computational Linguistics
  and the 7th International Joint Conference on Natural Language Processing of
  the Asian Federation of Natural Language Processing {ACL}}, vol.~2, July
  2015, pp. 495--500.

\bibitem{Mahoney2011}
\BIBentryALTinterwordspacing
M.~Mahoney, ``Large text compression benchmark,'' 2011. [Online]. Available:
  \url{http://mattmahoney.net/dc/textdata.html}
\BIBentrySTDinterwordspacing

\bibitem{Pascanu2013}
R.~Pascanu, {\c{C}}.~G{\"{u}}l{\c{c}}ehre, K.~Cho, and Y.~Bengio, ``How to
  construct deep recurrent neural networks,'' \emph{CoRR}, vol. abs/1312.6026,
  2013.

\bibitem{Glorot2010}
X.~Glorot and Y.~Bengio, ``Understanding the difficulty of training deep
  feedforward neural networks,'' in \emph{Proceedings of the Thirteenth
  International Conference on Artificial Intelligence and Statistics
  (AISTATS)}, Chia Laguna Resort, Sardinia, Italy, May 2010, pp. 249--256.

\bibitem{Mikolov2011b}
T.~Mikolov, A.~Deoras, D.~Povey, L.~Burget, and J.~Cernock{\'{y}}, ``Strategies
  for training large scale neural network language models,'' in \emph{{IEEE}
  Workshop on Automatic Speech Recognition {\&} Understanding (ASRU)},
  Waikoloa, HI, USA, Dec. 11-15, 2011, pp. 196--201.

\bibitem{Xu2007}
P.~Xu and F.~Jelinek, ``Random forests and the data sparseness problem in
  language modeling,'' \emph{Computer Speech {\&} Language}, vol.~21, no.~1,
  pp. 105--152, 2007.

\bibitem{Filimonov2009}
D.~Filimonov and M.~P. Harper, ``A joint language model with fine-grain
  syntactic tags,'' in \emph{Conference on Empirical Methods in Natural
  Language Processing (EMNLP), {A} meeting of SIGDAT, a Special Interest Group
  of the {ACL}}, Singapore, Aug. 2009, pp. 1114--1123.

\bibitem{Emami2004}
A.~Emami and F.~Jelinek, ``Exact training of a neural syntactic language
  model,'' in \emph{{IEEE} International Conference on Acoustics, Speech, and
  Signal Processing (ICASSP)}, Montreal, Quebec, Canada, May 2004, pp.
  245--248.

\bibitem{Mikolov2011c}
T.~Mikolov, A.~Deoras, S.~Kombrink, L.~Burget, and J.~Cernock{\'{y}},
  ``Empirical evaluation and combination of advanced language modeling
  techniques,'' in \emph{12th Annual Conference of the International Speech
  Communication Association (INTERSPEECH)}, Florence, Italy, Aug. 27-31, 2011,
  pp. 605--608.

\end{thebibliography}

\end{document}